\documentclass[11pt,a4paper]{article}
\usepackage[hyperref]{naaclhlt2018}
\usepackage{times}
\usepackage{latexsym}
\usepackage{graphicx}
\usepackage{url}

\aclfinalcopy 

\title{The Relevance of Text and Speech Features \\ in Automatic Non-native English Accent Identification}

\author{Sowmya Vajjala \\
 Iowa State University, USA\\
  {\tt sowmya@iastate.edu} \\ \And
 Ziwei Zhou \\
 Iowa State University, USA\\
{\tt ziweizh@iastate.edu} 
 }

\begin{document}
\maketitle
\begin{abstract}
This paper describes our experiments with automatically identifying native accents from speech samples of non-native English speakers using low level audio features, and n-gram features from manual transcriptions. Using a publicly available non-native speech corpus and simple audio feature representations that do not perform word/phoneme recognition, we show that it is possible to achieve close to 90\% classification accuracy for this task. While character n-grams perform similar to speech features, we show that speech features are not affected by prompt variation, whereas ngrams are. Since the approach followed can be easily adapted to any language provided we have enough training data, we believe these results will provide useful insights for the development of accent recognition systems and for the study of accents in the context of language learning.
\end{abstract}

\section{Introduction}
\label{intro}

Understanding and/or modeling native language (L1) influence on second (L2) language production has been a topic of research interest for a long time. Doing this with written language has several applications in domains such as customized language instruction \cite{Lu.Ai-15}, forensic linguistics, and stylistic studies \cite{Argamon.Koppel.ea-09}. Identifying L1 accent in L2 speech is particularly useful in applications such as personalized speech recognition and pronunciation tutoring \cite[e.g.,][]{Eskenazi.Kennedy.ea-07}. It is also an important challenge to address in the age when voice-driven interfaces are commonly used by speakers with diverse accents across the world \cite{Schuller.Steidl.ea-16}. Finally, understanding L1 influence on L2, whether in written or spoken language, is also useful in understanding the process of language learning. 

Considering these perspectives, there has been a surge in the research interest in this direction, as it can be seen from the recent Native Language Identification (NLI) shared tasks in the NLP community \cite{Tetreault.Blanchard.ea-13,Malmasi.Evanini.ea-17} and the Computational Paralinguistics Challenge in the Speech community \cite{Schuller.Steidl.ea-16}. There has been a lot of research into phoneme recognition based feature engineering for this task in recent past. Yet, the usefulness of low level audio features, and a comparison between audio and transcribed textual features have not been investigated systematically. Further, work on speech has been limited to proprietary datasets, often without access to the actual speech files, relying on intermediate representations. 

In this background, we investigate the usefulness of easy to extract text and audio features in identifying the accent in non-native speech using a publicly available dataset. We show that:
\begin{enumerate}\itemsep-1ex
\item audio features (without speech recognition) achieve close to 90\% classification accuracy in distinguishing between 10 Asian speech accents and native English speakers.
\item n-gram feature representations from manual transcriptions achieve comparable classification performance to these low-level speech features, but are sensitive to variations in the prompts/topics. 
\item audio features are not affected by prompt variation.
\end{enumerate} 

The rest of this paper is organized as follows. Section~\ref{sec:rel} briefly summarizes related work, and Sections~\ref{sec:methods} and \ref{sec:expts} describe our approach and results. Section~\ref{sec:rel} surveys the related work and Section~\ref{sec:conc} summarizes the main conclusions. 

\section{Related Work}
\label{sec:rel}
Automatic accent identification using non-native speech samples has seen a growing interest in the past few years\cite{Schuller.Steidl.ea-16,Malmasi.Evanini.ea-17} and the best performing systems used features based on i-vectors, and combinations of different word level features along with i-vectors. In comparison, there is not much work on how low level audio feature representations perform against features requiring speech recognition. 

Related research in NLI for written texts explored a range of languages, feature combinations and model ensembles in the past 5 years \cite[e.g.,][]{Tetreault.Blanchard.ea-13,Malmasi-16,Bich-17,Malmasi.Dras-17b}. In general, word and character n-gram features have given the best performing results for this task although these textual features were also shown to be sensitive to the training data \cite{Malmasi.Dras-15}. The NLI shared task in 2017 had both written and spoken tasks \cite{Malmasi.Evanini.ea-17}, in which manual transcriptions of speech files were provided along with i-vectors (instead of original audios). While i-vectors were shown to be useful for this task, the ngram features from manual transcriptions were shown to complement these features. However, manual transcriptions are unavailable in real world, beyond the experimental datasets. Yet, assuming the presence of automatic transcriptions will raise the question - "why can't we use ngram features from such transcriptions, as they perform the best with written language?" 

In this background, we explore the following questions in this paper:
\begin{enumerate}\itemsep-1ex
\item How far can we go without speech recognition/transcription for accent identification?
\item Can speech features work across prompts, unlike text ngram features?
\end{enumerate}

\section{Approach}
\label{sec:methods}
\paragraph{Corpus: } We used the speech part of the International Corpus Network of Asian Learners of English (ICNALE), which is a publicly and freely available corpus non-native writing and speech \cite{Ishikawa-14}. It contains 4400 English speech samples (approximately 1 minute in duration each), recorded in response to two prompts, along with plain text transcriptions. The data consists of speakers from 10 Asian countries and a sample of native English speakers. While there are other accent corpora such as the CSLU Foreign Accented English \cite{Lander-07} and the speech accents archive\footnote{\url{http://accent.gmu.edu/}}, we did not find them suitable for this task. The CSLU corpus had only one prompt, and no transcriptions, and the GMU corpus is not spontaneous speech. ICNALE corpus was used in the recent past \cite{Nisioi-16} in a similar task, to perform pair-wise NLI. The audio files in ICNALE underwent a morphing procedure for privacy reasons, to protect anonymity of participants. This involved altering the pitch and format to perform speaker normalization\footnote{The morphing software is publicly available on the corpus website}. Table~\ref{tab:corpus} shows the class distribution in the corpus. 

\begin{table}[htb!]
\begin{center}
\begin{tabular}{|l|l|}
\hline \bf Language & \bf Num. Audio files \\ \hline
Native English (ENS) & 600 \\
Hongkong English (HKG) & 200 \\
Pakistan (PAK) & 400 \\
Philippines (PHL) & 400 \\
Singpore (SIN) & 200 \\
China (CHN) & 600 \\
Indonesia (IDN) &400 \\
Japan (JPN) & 600 \\
Korea (KOR) & 400 \\
Thailand (THA) & 200 \\
Taiwan (TWN) & 400 \\ \hline
\end{tabular}
\end{center}
\caption{Composition of ICNALE spoken Corpus.}
\label{tab:corpus} 
\end{table}

\subsection{Features}
We used two kinds of feature representations, one for audio and one for transcriptions. For the audio features, we employed the low level acoustic descriptors baseline from INTERSPEECH ComParE challenges \cite{Schuller.Steidl.ea-13}, extracted using OpenSmile \cite{Eyben.Wollmer.ea-10}. This contains 6373 static features describing signal properties such as amplitude statistics, signal energy features, and features related to magnitude spectra, auto-correlation and cepstral characteristics \cite{Eyben-15}. These low-level features were known to be useful for performing a range of audio and music classification tasks in the past, and was also used as a baseline feature set in the first speech native language identification challenge. From the transcriptions, we extracted word and character n-gram features considering up to 3-grams for words and 10-grams for characters. We extracted character n-grams both with and without considering word boundaries. 

\subsection{Model selection and evaluation}
The dataset gives us a 11-class classification problem, and we explored a range of standard supervised learning algorithms for this purpose. Since the class distribution is unbalanced, we also experimented with oversampling the classes with less number of examples using SMOTE \cite{Chawla.Bowyer.ea-02} which creates additional synthetic examples for minority classes to balance the class distribution in training data. Model selection was done using cross-validation (CV) when the entire dataset is used, and by comparing results on test-set for cross prompt evaluation. Due to space constraints, we report only classification accuracy as our evaluation measure, as the manual inspection of confusion matrices did not show any apparent bias towards any class\footnote{Weighted F1 and other measures can be added, if necessary, in the final version of the paper}. The experiments were done using WEKA and scikit-learn \cite{Pedregosa.Varoquaux.ea-11,Buitinck.Louppe.ea-13}. 

\section{Experiments and Results}
\label{sec:expts}
\begin{figure}[htb!]
\centering
\includegraphics[width=0.4\textwidth,height=0.4\textheight,keepaspectratio]{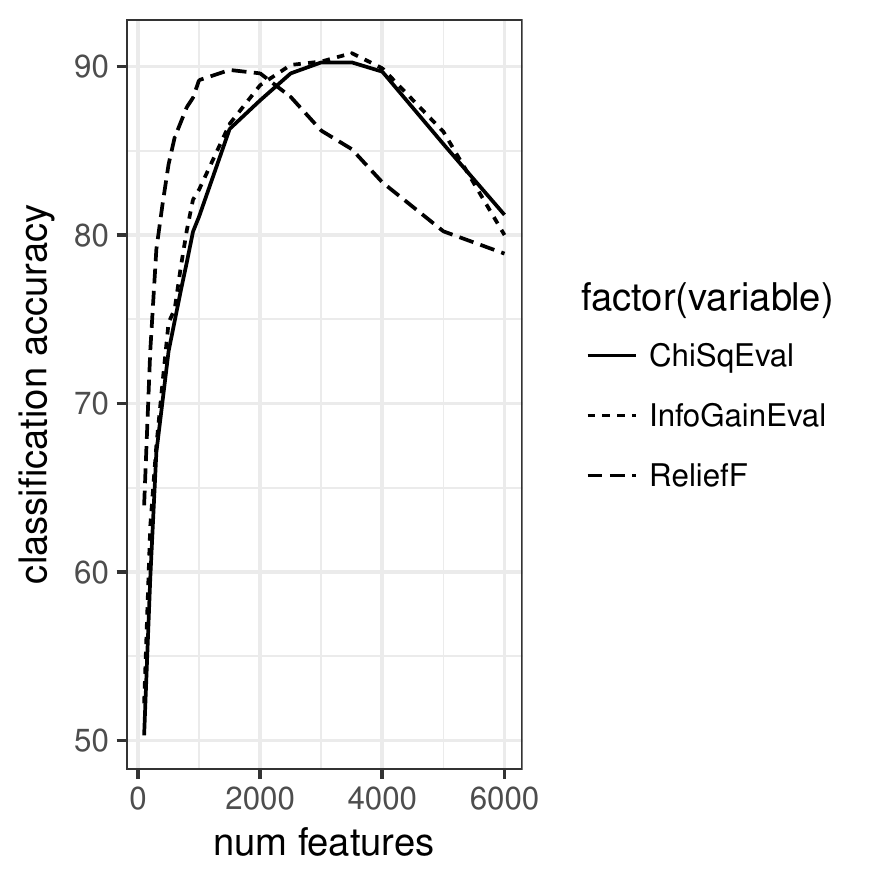}
\caption{Feature selection with speech features}
\label{fig:fs}
\end{figure}

\paragraph{With Speech Features: }
We trained classification models with all the 6373 audio features and with manual and automatic feature selection. Formant frequency features were shown to be useful for accent identification in early research \cite{Kat.Fung-99}. Others such as voicing features are meant to capture prosody in speech, and features based on MFCCs, and auditory spectrum features are used in speech recognition. So, these features can be considered as having some theoretical relevance for this task, compared to other features based on signal energy and other properties. Hence, we trained classifiers with these subsets. Table~\ref{tab:speechonly} shows the results for these experimental settings, with Sequential Minimal Optimization (SMO), which was the best performing classifier for these features in our experiments. 
\begin{table}[htb!]
\begin{center}
\begin{tabular}{| p{3cm}| l|p{1.35cm}|}
\hline \textbf{Description} & \textbf{num. feat.} &\textbf{Accuracy}\\
\hline all  (1) & 6373 &77.9\% \\
\hline Formant & 83& 45.2\%\\
\hline Voicing & 78& 42.0\%\\
\hline MFCCs&1400 & 69.4\%\\
\hline Aud. spec. & 100& 51.0\%\\
\hline \textbf{all feat. + SMOTE (2)} & 6373 & \textbf{90.8\%}\\
\hline
\end{tabular}
\end{center}
\caption{Accent identification with Speech features}
\label{tab:speechonly}  
\end{table}
The model with all the features, along with SMOTE oversampling, gives the best classification accuracy of 90.8\% for these features. Smaller feature subsets did not perform well, which indicates that the other audio signal features we excluded could be playing an important role in performing the classification. So, we explored automatic feature selection using three commonly used methods: Information gain, Chi-square, and ReliefF \cite{Kira.Rendell-92}, changing the number of top-N best features chosen (N=100 to 6000). Figure~\ref{fig:fs} shows a summary of these models, with cross-validation on the original training data (without SMOTE).

At around 3000 features, information gain and chi square selection methods seem to be giving the best performance, close to 90\%, and ReliefF reaches this performance with 2000 features. After that peak, results fall to around 80\% as we increase the features selected for all the three methods. While we did not look into the composition of these feature sets yet, it clearly shows even simple feature selection with baseline features result in large improvements in accuracies.

\paragraph{With Textual Features: }
Aside from speech based features, we trained classification models with word and character n-gram features from the transcriptions. Table~\ref{tab:textonly} summarizes the results with these features with Multinomial Logistic Regression (MLR), which gave the best results for these features. 
\begin{table}[htb!]
\begin{center}
\begin{tabular}{| p{4cm}| l|}
\hline \textbf{Features} & \textbf{Accuracy} \\
\hline Word n-grams (3) & 83.8\% \\
\hline Char n-grams across word-boundaries (4) & 88.3\% \\
\hline Char n-grams, w/o word boundaries (5) & 84.7\%\\ 
\hline (3) + (4) & 88.9\%\\
\hline 
\end{tabular}
\end{center}
\caption{Accent identification with n-grams from transcriptions}
\label{tab:textonly}  
\end{table}

From Table~\ref{tab:textonly}, we observe that n-gram features with minimal pre-processing seem to be extremely predictive of native accents. These results seem in contrast to the results from NLI Shared Task-2017 \cite{Malmasi.Evanini.ea-17}, where the word/character level features did not perform well as a stand-alone feature set with Speech data, but improved the performance when added to i-vector features, with the best performing system achieving an accuracy of 87.5\% combining i-vectors and transcription features. However, it is difficult to compare these results as they come from different datasets.

In our experiments, low level speech features are clearly doing well by themselves, achieving 90\% accuracy with feature selection. While they have only been used as baseline in contemporary approaches, there is also no systematic study on how far can we go with them without training speech recognition models. This paper shows that systematic feature selection may result in high accuracies for this task even with these baseline features.

\subsection{Prompt Specificity in Accent Identification}
In order to study the variation due to prompt, we split the dataset in to two parts. Since each participant responded to both the prompts in the corpus, the distribution of L1s in the training and test corpus remained the same. Table~\ref{tab:prompt} shows a summary of the results for different experimental settings (from Tables~\ref{tab:speechonly} and ~\ref{tab:textonly}), for 10-fold CV per prompt, and evaluation on the other prompt. The row \textbf{best} indicates the best performing feature configuration from Figure 1 (3500 features selected using information gain). As mentioned earlier, results for ngram features are with MLR and results for speech features are with SMO. Speech and text features were not combined in this paper as the text features are not shown to generalize across prompts. 

\begin{table}[htb!]
\begin{center}
\begin{tabular}{| p{1cm}| l| p{1.25cm} || l| p{1.25cm} |}
\hline set. & P1-CV & Train:P1, Test:P2 & P2-CV & Train:P2, Test:P1 \\
\hline 
\hline (1) &70.6\%& 74.9\%&71.3\% &72.8\%\\  
\hline (2) & 88.4\%& 75.3\%&88.2\%& 73.04\%\\
\hline (3) &83.1\% &52.6\% &81.95\% & 41.8\%\\ 
\hline (4)  & 88.2\%& 58.2\%&87.95\% &57.6\%\\ 
\hline (5)  & 85.4\%& 44.95\%& 82.95\%& 38.3\%\\ 
\hline (3)+(4) &  88.81\%&56.1\%& 88\%& 52.2\% \\
\hline \textbf{best} & \textbf{83.1\%}& \textbf{87.3\%}&\textbf{83.9\%} &\textbf{85.8\%}\\
\hline 
\end{tabular}
\end{center}
\caption{Accuracy for prompt specific experiments}
\label{tab:prompt} 
\end{table}
Clearly, there is a huge drop in accuracy from one prompt to another for the n-gram features, sometimes as much as 40\%. While over-sampling resulted in a increased classification accuracy in the prompt specific evaluation for speech features, it did not do improve cross-prompt evaluation. Cross-prompt evaluation results being better than same prompt evaluation for (1) and \textit{best} settings is primarily due to the fact that there is a +/- 5\% variation between CV folds, and we report only the average. Overall, these results lead us to a conclusion that employing low level audio features, without speech recognition and without transcriptions can possibly achieve generalizable speech accent identification.

\section{Conclusion}
\label{sec:conc}
Our experiments show that speech features from low level audio representations achieve over 90\% classification accuracy for a 11-class native accent identification problem, after performing feature selection. Further, the results indicate that these features can be potentially prompt independent, which has been a consistent issue regarding the generalizability of NLI models in the past. These encouraging results lead us to several interesting problems to explore in accent identification. Immediate extensions include comparing this with i-vector features, and exploring the usefulness of neural network architectures for the task. Additionally, since the approach does not have language specific components in the pipeline, we plan to replicate the experiments with L2 German and Arabic for which such corpora are freely available. Another interesting dimension to explore is to use text features from automatic transcriptions, as the transcription errors perhaps capture accent differences while manual transcriptions cannot. 

\bibliographystyle{acl_natbib}
\bibliography{coling2018}

\begin{thebibliography}{}
\expandafter\ifx\csname natexlab\endcsname\relax\def\natexlab#1{#1}\fi

\bibitem[{Argamon et~al.(2009)Argamon, Koppel, Pennebaker, and
  Schler}]{Argamon.Koppel.ea-09}
Shlomo Argamon, Moshe Koppel, James~W Pennebaker, and Jonathan Schler. 2009.
\newblock Automatically profiling the author of an anonymous text.
\newblock {\em Communications of the ACM\/} 52(2):119--123.

\bibitem[{Bich(2017)}]{Bich-17}
Serhiy Bich. 2017.
\newblock {\em Is There Choice in Non-Native Voice?" Linguistic Feature
  Engineering and a Variationist Perspective in Automatic Native Language
  Identification\/}.
\newblock Ph.D. thesis, University of T\"ubingen, Germany.

\bibitem[{Buitinck et~al.(2013)Buitinck, Louppe, Blondel, Pedregosa, Mueller,
  Grisel, Niculae, Prettenhofer, Gramfort, Grobler, Layton, VanderPlas, Joly,
  Holt, and Varoquaux}]{Buitinck.Louppe.ea-13}
Lars Buitinck, Gilles Louppe, Mathieu Blondel, Fabian Pedregosa, Andreas
  Mueller, Olivier Grisel, Vlad Niculae, Peter Prettenhofer, Alexandre
  Gramfort, Jaques Grobler, Robert Layton, Jake VanderPlas, Arnaud Joly, Brian
  Holt, and Ga{\"{e}}l Varoquaux. 2013.
\newblock {API} design for machine learning software: experiences from the
  scikit-learn project.
\newblock In {\em ECML PKDD Workshop: Languages for Data Mining and Machine
  Learning\/}. pages 108--122.

\bibitem[{Chawla et~al.(2002)Chawla, Bowyer, Hall, and
  Kegelmeyer}]{Chawla.Bowyer.ea-02}
Nitesh~V Chawla, Kevin~W Bowyer, Lawrence~O Hall, and W~Philip Kegelmeyer.
  2002.
\newblock Smote: synthetic minority over-sampling technique.
\newblock {\em Journal of artificial intelligence research\/} 16:321--357.

\bibitem[{Eskenazi et~al.(2007)Eskenazi, Kennedy, Ketchum, Olszewski, and
  Pelton}]{Eskenazi.Kennedy.ea-07}
Maxine Eskenazi, Angela Kennedy, Carlton Ketchum, Robert Olszewski, and Garrett
  Pelton. 2007.
\newblock The nativeaccenttm pronunciation tutor: measuring success in the real
  world.
\newblock In {\em Workshop on Speech and Language Technology in Education\/}.

\bibitem[{Eyben(2015)}]{Eyben-15}
Florian Eyben. 2015.
\newblock {\em Real-time speech and music classification by large audio feature
  space extraction\/}.
\newblock Springer.

\bibitem[{Eyben et~al.(2010)Eyben, W{\"o}llmer, and
  Schuller}]{Eyben.Wollmer.ea-10}
Florian Eyben, Martin W{\"o}llmer, and Bj{\"o}rn Schuller. 2010.
\newblock Opensmile: the munich versatile and fast open-source audio feature
  extractor.
\newblock In {\em Proceedings of the 18th ACM international conference on
  Multimedia\/}. ACM, pages 1459--1462.

\bibitem[{Ishikawa(2014)}]{Ishikawa-14}
Shin'ichiro Ishikawa. 2014.
\newblock Design of the icnale-spoken: A new database for multi-modal
  contrastive interlanguage analysis.
\newblock {\em Learner corpus studies in Asia and the world\/} 2:63--76.

\bibitem[{Kat and Fung(1999)}]{Kat.Fung-99}
Liu~Wai Kat and Pascale Fung. 1999.
\newblock Fast accent identification and accented speech recognition.
\newblock In {\em Acoustics, Speech, and Signal Processing, 1999. Proceedings.,
  1999 IEEE International Conference on\/}. IEEE, volume~1, pages 221--224.

\bibitem[{Kira and Rendell(1992)}]{Kira.Rendell-92}
Kenji Kira and Larry~A Rendell. 1992.
\newblock The feature selection problem: Traditional methods and a new
  algorithm.
\newblock In {\em Aaai\/}. volume~2, pages 129--134.

\bibitem[{Lander(2007)}]{Lander-07}
Terri Lander. 2007.
\newblock Cslu: Foreign accented english release 1.2.
\newblock {\em Linguistic Data Consortium, Philadelphia\/} .

\bibitem[{Lu and Ai(2015)}]{Lu.Ai-15}
Xiaofei Lu and Haiyang Ai. 2015.
\newblock Syntactic complexity in college-level english writing: Differences
  among writers with diverse l1 backgrounds.
\newblock {\em Journal of Second Language Writing\/} 29:16--27.

\bibitem[{Malmasi(2016)}]{Malmasi-16}
Shervin Malmasi. 2016.
\newblock {\em Native language identification: explorations and
  applications\/}.
\newblock Ph.D. thesis, PhD Thesis, Macquarie University.

\bibitem[{Malmasi and Dras(2015)}]{Malmasi.Dras-15}
Shervin Malmasi and Mark Dras. 2015.
\newblock Large-scale native language identification with cross-corpus
  evaluation.
\newblock In {\em HLT-NAACL\/}. pages 1403--1409.

\bibitem[{Malmasi and Dras(2017)}]{Malmasi.Dras-17b}
Shervin Malmasi and Mark Dras. 2017.
\newblock Multilingual native language identification.
\newblock {\em Natural Language Engineering\/} 23(2):163--215.

\bibitem[{Malmasi et~al.(2017)Malmasi, Evanini, Cahill, Tetreault, Pugh,
  Hamill, Napolitano, and Qian}]{Malmasi.Evanini.ea-17}
Shervin Malmasi, Keelan Evanini, Aoife Cahill, Joel Tetreault, Robert Pugh,
  Christopher Hamill, Diane Napolitano, and Yao Qian. 2017.
\newblock A report on the 2017 native language identification shared task.
\newblock In {\em Proceedings of the 12th Workshop on Innovative Use of NLP for
  Building Educational Applications\/}. pages 62--75.

\bibitem[{Nisioi(2016)}]{Nisioi-16}
Sergiu Nisioi. 2016.
\newblock Comparing speech and text classification on icnale.
\newblock In Nicoletta Calzolari~(Conference Chair), Khalid Choukri, Thierry
  Declerck, Sara Goggi, Marko Grobelnik, Bente Maegaard, Joseph Mariani, Helene
  Mazo, Asuncion Moreno, Jan Odijk, and Stelios Piperidis, editors, {\em
  Proceedings of the Tenth International Conference on Language Resources and
  Evaluation (LREC 2016)\/}. European Language Resources Association (ELRA),
  Paris, France.

\bibitem[{Pedregosa et~al.(2011)Pedregosa, Varoquaux, Gramfort, Michel,
  Thirion, Grisel, Blondel, Prettenhofer, Weiss, Dubourg, Vanderplas, Passos,
  Cournapeau, Brucher, Perrot, and Duchesnay}]{Pedregosa.Varoquaux.ea-11}
F.~Pedregosa, G.~Varoquaux, A.~Gramfort, V.~Michel, B.~Thirion, O.~Grisel,
  M.~Blondel, P.~Prettenhofer, R.~Weiss, V.~Dubourg, J.~Vanderplas, A.~Passos,
  D.~Cournapeau, M.~Brucher, M.~Perrot, and E.~Duchesnay. 2011.
\newblock Scikit-learn: Machine learning in {P}ython.
\newblock {\em Journal of Machine Learning Research\/} 12:2825--2830.

\bibitem[{Schuller et~al.(2013)Schuller, Steidl, Batliner, Vinciarelli,
  Scherer, Ringeval, Chetouani, Weninger, Eyben, Marchi
  et~al.}]{Schuller.Steidl.ea-13}
Bj{\"o}rn Schuller, Stefan Steidl, Anton Batliner, Alessandro Vinciarelli,
  Klaus Scherer, Fabien Ringeval, Mohamed Chetouani, Felix Weninger, Florian
  Eyben, Erik Marchi, et~al. 2013.
\newblock The interspeech 2013 computational paralinguistics challenge: social
  signals, conflict, emotion, autism.
\newblock In {\em Proceedings INTERSPEECH 2013, 14th Annual Conference of the
  International Speech Communication Association, Lyon, France\/}.

\bibitem[{Schuller et~al.(2016)Schuller, Steidl, Batliner, Hirschberg, Burgoon,
  Baird, Elkins, Zhang, Coutinho, and Evanini}]{Schuller.Steidl.ea-16}
Bj{\"o}rn~W Schuller, Stefan Steidl, Anton Batliner, Julia Hirschberg, Judee~K
  Burgoon, Alice Baird, Aaron~C Elkins, Yue Zhang, Eduardo Coutinho, and Keelan
  Evanini. 2016.
\newblock The interspeech 2016 computational paralinguistics challenge:
  Deception, sincerity \& native language.
\newblock In {\em INTERSPEECH\/}. pages 2001--2005.

\bibitem[{Tetreault et~al.(2013)Tetreault, Blanchard, and
  Cahill}]{Tetreault.Blanchard.ea-13}
Joel~R Tetreault, Daniel Blanchard, and Aoife Cahill. 2013.
\newblock A report on the first native language identification shared task.
\newblock In {\em BEA@ NAACL-HLT\/}. pages 48--57.

\end{thebibliography}

\end{document}